# Machine learning in bioprocess development: From promise to practice

Laura M. Helleckes[1,2], Johannes Hemmerich[1,#], Wolfgang Wiechert[1,2], Eric von Lieres[1,2], and Alexander Grünberger [3,4*]

[1] *Institute for Bio- and Geosciences (IBG-1), Forschungszentrum Jülich GmbH, 52428 Jülich, Germany*

[2] *RWTH Aachen University, Templergraben 55, 52062 Aachen*

[3] *Multiscale Bioengineering, Technical Faculty, Bielefeld University, Universitätsstr. 25, 33615 Bielefeld, Germany*

[4] *Center for Biotechnology (CeBiTec), Bielefeld University, Universitätsstr. 25, 33615 Bielefeld, Germany*

[#] *Current address: LenioBio GmbH, Technology & Innovation Center, Forckenbeckstr. 6, 52074 Aachen, Germany*

**Correspondence:** alexander.gruenberger@uni-bielefeld.de

## Abstract

Fostered by novel analytical techniques, digitalization and automation, modern bioprocess development provides high amounts of heterogeneous experimental data, containing valuable process information. In this context, data-driven methods like machine learning (ML) approaches have a high potential to rationally explore large design spaces while exploiting experimental facilities most efficiently. The aim of this review is to demonstrate how ML methods have been applied so far in bioprocess development, especially in strain engineering and selection, bioprocess optimization, scale-up, monitoring and control of bioprocesses. For each topic, we will highlight successful application cases, current challenges and point out domains that can potentially benefit from technology transfer and further progress in the field of ML.

**Machine learning in biotechnology: State of the art**

Over the past years, machine learning (ML) has become the most important discipline of artificial intelligence (AI) in terms of practical application. ML deals with algorithms and programs that learn to solve certain tasks based on data, where performance increases with experience, i.e. available data [1]. More precisely, ML aims at finding suitable, mostly empirical models to describe datasets, learning from labeled samples or by identifying inherent patterns (see **Box 1** for central paradigms). The vast spectrum of ML methods **[Box 3]** is particularly useful when large amounts of data are available and/or when datasets are too complex to be analyzed by sets of predefined rules (see the explanation of expert systems in **Box 1)**. Other applications of ML aim at finding so-called **surrogate models**, in which ML models are used as approximations for mechanistic models that are costly or hard to evaluate [2].

In previous years, life sciences have started looking into available ML methods and researchers began to assess which of these methods are suitable to tackle current challenges [3]. Thus, biology and biotechnology became influenced by recent advances in ML. This is reflected by many reviews, for example ML in protein function prediction [4], multi-omics data analysis [5], developmental biology [6], biological network analysis [7], metabolic engineering [8] and biochemical engineering [9].

Generally, the biotechnological pipeline from target molecule to final product covers four essential stages, which are (i) target identification and molecule design, (ii) biocatalyst design, (iii) bioprocess development as well as (iv) industrial-scale production. The first two stages are mainly addressed by molecular biotechnology and bioinformatics; in recent years, both fields were heavily influenced by technological progress on the experimental side (e.g., omics technologies) as well as increased computational power [10]. The resulting availability of big data and compute resources enabled the rise of ML, which nowadays is state-of-the-art. A notable, recent breakthrough of ML is AlphaFold [11], a deep learning **[Box 2]** program that predicts the 3D structure of proteins from sequence data. Since ML is abundant and diverse for the first two stages, i.e., molecule prediction and biocatalyst design, a thorough review is out-of-scope for this paper. The reader is instead referred to existing reviews, e.g., [7,12-15].

The third stage of the biotechnological production pipeline, **bioprocess development**, focuses on increasing the production capacity for the target molecule by means of strain selection, process optimization, and scale-up. During this stage, high-throughput screening (HTS) experiments are typically performed to assess the performance of selected clones [16]. Furthermore, optimal cultivation parameters need to be identified from a huge design space. However, traditional analytical methods will often not match the rate of experimentation and thus, analysis subsequently becomes a bottleneck [17,18]. To overcome this challenge, smart experimental design and accompanying modeling or ML are needed to maximize information content of the designed experiments.

Since biological and process parameters are correlated, iterative experimentation and data evaluation is needed to feed back information and insights from screening to strain design. This approach is reflected in the **Design-Build-Test-Learn (DBTL)** cycle [19], which is sometimes only referring to synthetic biology, but can be also applied to the stage of bioprocess development [20]. In the context of DBTL, the learning step can be enhanced by ML, which can suggest smart designs for the next round of experimentation [21].

The fourth and final stage of the biotechnological production pipeline is concerned with reproducible and robust operation, for example by controlling raw materials [22] of industrial-scale bioprocesses. Research at this stage is determining the long-term stable and consistent operation of a production process, for example by process intensification at large scale to increase manufacturing capacity [23]. Methods and results at this stage are often proprietary and therefore scarcely available in public literature [24]. In market and business analyses, necessary productivity ranges are determined to meet bioprocess economics, a prerequisite for the fourth stage.

In the light of increased automation, data availability and exchange, data-driven bioprocess development will accelerate the time-to-market of bioproducts [25]. Moreover, the large cash flow in the AI sector will boost the integration of novel methods to the development pipeline [24]. In this review we will thus mainly discuss the application of ML in bioprocess development, particularly in upstream processes. ML is also advancing in downstream processing, where corresponding techniques are developed for specific technologies such as chromatography [26-31], but also for complex purification pipelines of specific products such as antibodies [32] or inclusion bodies [33]. Since available literature is highly diverse and vast enough to be covered in a separate review, ML for downstream processing is not discussed in detail here. Where possible, ML trends in commercial bioprocesses, e.g. for **Process Analytical Technology (PAT)**, digital twins or model predictive control (MPC**)**, are included. Overall, four main topics are addressed:

1. strain selection and engineering
2. bioprocess optimization
3. scale-up of bioprocesses
4. process monitoring and control.

For each area, we shed light on methodological milestones of the past years that push ML forward. Furthermore, we discuss potentials and challenges for future usage and improvement of ML tools. This leads to an overall discussion about ML as an enabling technology for bioprocess development.

**Choosing between numerous candidates: strain engineering and selection**

One central step before bioprocess development takes place is the selection of a biocatalyst or microorganism for production. State-of-the-art experimental methods for high-throughput screening exist to identify potent biocatalysts, e.g. by quantitative phenotyping of strain libraries [16,34,35]. The current bottleneck is thus automated data processing and algorithm-driven decision making to select biocatalysts with the highest potential for commercial production.

Recent advances in ML provide a number of techniques to foster biochemical engineering of strains [8]. As a major challenge, the diversity of biocatalysts leads to a broad range of possible tasks, for example design and selection of bacterial production strains, predicting production in different cell-free systems or engineering mammalian cell lines. The latter poses many additional challenges such as clonal variation [36], and large-scale studies are needed to generate mechanistic understanding, which is so far required for non-ML methods [37]. To maintain focus, we review strain selection here. For insights into ML approaches for enzyme and biocatalyst engineering, the reader is referred to [38,39].

In the past decades, stoichiometric as well as kinetic genome-scale models have been used for both metabolic engineering and bioprocess development [40-42]. Besides genetic design, such models can give insights into suitable carbon sources, media design or bioreactor parameters [43]. For many years, quantitative predictions for metabolic engineering have been made using **constraint-based modeling (COBRA)** of genome-scale metabolic networks [44,45]. Methods of the COBRA toolbox like Flux Balance Analysis (FBA) [46], Minimization of Metabolic Adjustment (MOMA) [47], or Minimal Cut Sets (MCS) [48] generally aim to optimize fluxes in a biological network (i.e. metabolism) to improve productivity by, e.g. reducing side product formation or eliminating competing metabolic pathways. Resolving metabolic pathways and determining corresponding fluxes is experimentally demanding. Within the COBRA toolbox, FBA is probably the most popular method to find steady-state flux solutions. However, FBA is largely limited by understanding of the underlying network structure [45].

In contrast, data-driven ML algorithms allow for analysis of large, complex (multi-)omics datasets, which can be generated in high throughput [10,49]. Different applications of ML for genome-scale models are emerging. On the one hand, ML is used to complement the typical modeling pipeline for constraint-based models, namely in the steps of gene annotation, gap filling and integration of multi-omics data [50]. On the other hand, novel approaches of **hybrid modeling** have been proposed, as well as ML methods that replace the mechanistic, genome-scale models completely. King *et al.* applied literature mining [**Box 3**] to create a database of *Escherichia coli* strain variants and their byproduct streams. They demonstrated how the database can be used to validate different genome-scale models [51]. Oyetunde *et al.* [52] combined simulated data from a genome-scale model with a manually curated dataset of bioprocess data from different *E. coli* strains as input data to predict production metrics for various products. As ML methods, they applied a combination of principal component analysis (PCA) and ensemble learning [**Box 2**], a strategy where different ML algorithms are combined to learn more efficiently (**Figure 1A**). The approach led to decent predictions of production titers, rates and yields (TRY) under varying process and pathway conditions, thus demonstrating the potential of integrating both large datasets and mechanistic knowledge from the genome-scale model.

In a similar approach, Zhang *et al.* [53] first used a genome-scale model to identify relevant genes for metabolic engineering of tryptophan production strains to then apply ensemble learning on biosensor data generated with promoter libraries of the suggested genes. In contrast to Oyetunde *et al.* [52] this model does not use the genome-scale model predictions as training data, but only to identify genomic targets for the promoter libraries. The developed ML models were used to predict combinations of promoters and genes outside the training dataset, thus augmenting the experimentally tested designs. Although this approach led to even further improved variants, the authors observed a reduced performance of the algorithm regarding extrapolation, which is a commonly known problem of ML approaches.

Finally, a recent preprint introduced Artificial Metabolic Networks (AMN) [54], a concept where fluxes are predicted with a recurrent neural network (RNN) [**Box 3**]. Here, FBA predictions are used to train the AMN, which can in turn replace the genome-scale model in the application phase. Since the AMN allows for **backpropagation,** it can be used to predict uptake rates based on external concentrations. Further methods of combining ML with genome-scale metabolic models, e.g. fluxomic analysis, have been recently reviewed [50,55,56].

The mentioned study of Zhang *et al*. [53] made use of the recently developed Automated Recommendation Tool for Synthetic biology [57]. The toolbox combines the Scikit-learn framework [58] with a **Bayesian statistics** approach of ensemble models, i.e. several models are trained for prediction and to provide uncertainty quantification. Moreover, the toolbox is flexible w.r.t. experimental requirements such as with the DBTL cycle, for which it can provide recommendations on strain design for the next iteration. This tool is an interesting starting point to further apply ML for synthetic biology [57,59] and should be expanded by the community.

Regarding kinetic models, detailed genome-scale models are often under-determined, meaning that the large number of kinetic parameters cannot be estimated from experimental data [60]. A remaining challenge thus is that these under-determined mathematical systems allow a multitude of parameter combinations that can equally well describe experimental measurements. However, many frameworks that determine the spaces of possible parameters propose a number of models that inherently contradict the experimentally observed physiology. To overcome this challenge, the REKINDLE (REconstruction of KINetic models using Deep LEarning) framework was recently suggested, in which Generative Adversarial Networks (GANs) [**Box 3**] are used to obtain mechanistic, kinetic models with biologically feasible dynamics [61].

Guiding strain optimization, Sabzevari *et al*. [62] applied a multi-agent reinforcement learning algorithm to both experimental data and data from a genome-scale kinetic model to tune metabolic enzyme levels. The algorithm outperforms another ML approach, namely Bayesian optimization on Gaussian processes (GPs) [**Box 3**], as well as a random search approach. Moreover, the multi-agent reinforcement learning approach allows for including parallelized experiments, which is important to make efficient use of modern HTS.

Overall, most studies on strain engineering and selection so far only focus on model host organisms and only few studies looked into transferability to other host systems [63]. A promising tool in this context could be transfer learning [**Box 2**]; however, this approach usually requires large amounts of data to train the initial model [9], which are frequently not available. In the field of strain modeling, transfer learning is still underexplored. We thus identify predictions for non-model organisms as well as providing comparable results for a variety of different strains as major bottlenecks. Most likely, no single ML algorithm will solve this issue; instead, a suite of ML algorithms is needed to cover such a complex task in the future.

**Raising and stabilizing titers, rates and yields (TRY): bioprocess optimization**

During bioprocess development and optimization, lab-scale bioprocesses are used to improve TRY by identifying optimal physico-chemical parameters for cultivation. In this context, different ML techniques are used.

Aiming at the application of microorganisms and enzymes at extreme temperatures, Li *et al*. [64] (**Figure 1B**) developed a support vector machine (SVM) [**Box 3**] regression model to predict the optimal temperature for enzymatic activity, using optimal growth temperature and amino acid sequence information as input features. Another common ML approach for bioprocess optimization is GP regression. Use-cases include optimization of pigment production in algae [65,66] and tuning of media composition for protein production in *Corynebacterium glutamicum* [67].

Finally, Artificial Neural Networks (ANNs) [**Box 3**] are frequently applied for a range of applications, e.g. optimizing media composition for wheat germ [68] or for pigment production in cyanobacteria [69]. Other studies optimize fermentation parameters; Pappu *et al*. [70], for example, investigated temperature, fermentation time, pH, $k_L a$, biomass and glycerol as influential parameters for xylitol production in the yeast *Debaryomyces nepalensis*. Ebrahimpour *et al*. [71] optimized production of a thermostable lipase in a *Geobacillus* strain with growth temperature, medium volume, inoculum size, agitation rate, incubation period and initial pH as input variables. Finally, some studies look into the complex interaction of media composition and fermentation parameters, e.g. in bioethanol production with *Saccharomyces cerevisiae* [72] or growth of cell lines for therapy [73].

Aiming at the transfer of knowledge between different bioprocesses, Rogers *et al*. simulated dynamical behavior in biochemical processes for different organisms via transfer learning, in this case by partially preserving layers between different ANNs [74]. Hutter and coworkers [75] combined GP regression with transfer learning, more precisely embedding vectors [**Box 3**], a technique that is used in **Natural Language Processing** to quantify similarity between words [76]. Both approaches show how historic data can be used to predict dynamics for new products, which is beneficial for bioprocess optimization.

Video and image data, e.g., of cell morphology, is a rich source of information for bioprocess analysis and control [77,78]. Here, microfluidic systems in combination with life-cell imaging have been pioneering the image analysis methods, among others for HTS of strains and to get improved understanding of cellular behavior at bioprocess-relevant cultivation conditions [79,80]. Deep learning techniques [**Box 2**] are well-suited to process such complex raw data from images in an automated fashion, thus laying the foundation for microfluidic-assisted, high-throughput bioprocess development [81,82]. Recent examples include prediction of growth and dynamics in microfluidic single-cell cultivation [83,84] and microfluidic droplet reactors, where multi-layer ANNs were used to predict performance of flow-focusing droplet generators [85].

Other applications in process optimization include the use of microscopic image data for spatio-temporal analysis of biofilms [86] and algae cultivation [87]. The latter requires complex management of light conditions and growth patterns, e.g. to avoid mutual shading during cultivation [88,89]. Here, Long *et al*. [87] used SVM regression to predict light distribution patterns from microscopy images, which provide insight into the growth behavior and could ultimately help to develop novel cultivation designs.

Finally, we see the advance of ML in automated flowsheet synthesis in chemical engineering [90,91]. Although not yet demonstrated for bioprocesses, such techniques have great potential to accelerate bioprocess development.

**Challenges at the verge to commercial scale: bioprocess scale-up**

Having selected strains and optimized process conditions at laboratory scale, a bioprocess needs to be transferred to industrial production scale. The production scale is typically subject to increased variability of materials and feed streams, more complex hydrodynamics and decreased spatial homogeneity [92]. Physical scale-down simulators, i.e., networks of purposefully heterogeneous laboratory devices, can be used to select strains and optimize process conditions under industrially relevant conditions [92,93]. Detailed measurements in industrial equipment are often intricate and prohibitively expensive. Model-based scale-up and scale-down simulators can help with closing this gap [93,94] and ensure industrial feasibility of the designed processes [95]. These models can facilitate the transfer of scale-independent knowledge and information, mostly related to the catalyst, while correcting the influence of scale-dependent mechanisms, mostly related to transport phenomena. Certain challenges arise when ML models, e.g., of cell metabolism, that have been trained at laboratory scale need to extrapolate beyond the experienced environmental conditions when applied at production scale.

Scale-up modeling involves various interconnected relationships between bioreactor design parameters, process parameters and hydrodynamic characteristics, which are almost impossible to theoretically describe and computationally trace using mechanistic models [96]. Scale-up has thus been studied with multivariate data analysis, where methods such as PCA, SVM regression and PLS are overlapping with ML, e.g., [97,98]. Recent ML approaches such as RNNs [99] or decision trees [100] as well as hybrid models [101] are starting to emerge for incorporation and transfer of information between different scales.

In their study, Bayer *et al*. [101] investigated hybrid models (differential equations/ANNs) for **Design of Experiments (DoE)** and **intensified DoE (iDoE)** experiments with mammalian cells at different scales; more precisely, experiments were run in shaker-scale, bolus fed-batch experiments as well as continuously fed 15L scale bioreactors. The authors found that a hybrid model trained on shaker-scale DoE data performed well on test data of the 15L reactor, especially for the iDoE data (**Figure 1C**). Moreover, a second model was trained on 15L iDoE experiments, in which intra-experiments variations are made. This model also performed reasonably well on data from 15L static runs (without variation), indicating that the iDoE concept can be generalized for modeling of mammalian cells. Compared to other studies, which often retrain the whole model for different scales, thus requiring large amounts of data, the results of Bayer *et al*. [101] are promising since they show the possibility of generalizing scale-up models across scales, meaning that a model trained for small-scale is still valid at large scale without retraining.

Overall, we consider ML to have great potential in the discovery of non-traditional scale-up criteria, for example by correlating validated mechanistic models describing bioprocess performance at laboratory and production scales. Beyond this application, ML models can also be used as surrogate models for complex scale-up models, e.g. by replacing costly simulations in computational fluid dynamics [102]. Though literature in the field of ML learning for bioprocess scale-up is still scarce, we anticipate that methods will be evolving quickly, potentially using the field of chemical engineering as a blueprint, e.g., [103].

**Monitoring and controlling bioprocesses: ML in Process Analytical Technology (PAT)**

In the final phases of bioprocess development, the transition to commercial production is targeted. Here, process monitoring and control are important steps, especially if the desired product needs to meet complex (pharmaceutical) regulations by the Food and Drug Administration (FDA) or European Medicines Agency (EMA) [104]. The need to record process data in a structured way to prove consistency for regulatory approval offers the opportunity to apply ML techniques. To support effective and efficient monitoring of **Critical Process Parameters (CPPs)** in bio(pharmaceutical) processes, the FDA introduced the **PAT** initiative [105]. Since controlling CPPs is pivotal to ensure validated ranges of **Critical Quality Attributes (CQAs)** of the product, PAT aims to establish regulatory approved monitoring capabilities for in-process controls, thus ensuring sufficient quality of the final product while the bioprocess is running.

**Soft sensors**, which are of high interest in the PAT framework [106], use mathematical models (<u>soft</u>ware) to make real-time predictions of a system, similar to hardware <u>sensors</u> [107]. Soft sensors can provide information about process variables that cannot be measured reliably or at all by mapping their prediction to frequent online data [108]; the corresponding models have to be constantly updated to fit the online process data best. Soft sensor models are structured in three different classes as well as any combination thereof: mechanistic models, multivariate statistics and AI/ ML [109]. In the context of this review, we focus on the latter; however, several general reviews on soft sensors for bioprocessing exist [104,110,111]. Soft sensors are also important building blocks for digital twins, which predictively describe the production process behavior [112,113]. A key feature of digital twins is bi-directional data exchange between a physical process and its model twin [114]. By analyzing the systems behavior *in silico*, further experimentation is guided efficiently towards process validation and qualification since the corresponding DBTL cycle iterations can be run faster [115,116].

Many ML-based approaches for soft sensors rely on ANNs [**Box 3**] or SVMs [**Box 3**] [117] [**Table 1**]. Successful examples include ANN-based soft sensors for erythromycin production [118] or biomass estimation in plant cell cultures [119] as well as the description of L-lysine fermentation process data using a multi-output least squares SVM regressor [**Box 3**] [120]. Recent advances in the field make use of deep learning [**Box 2**] instead [121]. For example, Gopakumar *et al*. [122] demonstrated that their deep soft sensor outperforms traditional SVM approaches for non-linear systems, shown for crucial parameters in two fermentation processes (Streptokinase and Penicillin). Interestingly, Yao *et al*. [123] developed a soft sensor that combines unsupervised learning for feature extraction with a semisupervised classification approach. Finally, Mowbray *et al*. [124] incorporated uncertainty quantification and non-linearities in their soft-sensors by using probabilistic ML methods such as Bayesian neural networks [**Box 3**]. All three studies highlight the potential of transferring ML technologies to the field of process monitoring to outperform conventional approaches.

However, similar to other applications, particularly in process scale-up, ML approaches for soft sensors suffer from the problem of transferability. In particular, the training dataset and the actual system variables have to share the same feature space [125], making it hard to transfer models to new plants. Moreover, older plants are prone to changing process conditions, which require adaptation in the models as well [126]. These challenges require novel techniques of transfer learning, which are currently at the starting point of their application in bioprocessing [125]. We believe that adaptation to new use-cases will be crucial for ML techniques to truly enhance process monitoring in biotechnology.

As an example from chemical engineering, Li *et al*. [127] implemented fault detection for a continuous stirred tank reactor and a plant-wide pulp mill by means of deep and transfer learning. Using simulated instead of measured data to train an ML algorithm inevitably leads to model-process mismatch since no mechanistic model can perfectly describe the real process. To overcome this challenge and use simulated data to train their convolutional neural network (CNN) [**Box 3**], the authors applied transfer learning for domain adaptation, meaning that measured data from other processes is used as well to increase prediction performance. Similar approaches could also enhance process monitoring in biotechnology.

Towards controlling bioprocesses, ML can have a high impact regarding model predictive control (MPC) [**Box 3**]. In principle, MPC is a methodology that uses three components: a model to predict system outputs, an objective function as well as a control law [128]. As an advantage, MPC can optimize performance of a system while considering constraints [129]. Here, ML is particularly useful for complex non-linear systems or systems for which little process understanding exists [130].

Nagy [131] used a detailed, mechanistic process model of a yeast fermentation, including biomass, media concentration and oxygen, to generate training data for an ANN. The ANN was then shown to efficiently replace the mechanistic model in MPC. Masampally [132] implemented a cascade structure of GP regression submodels, which can predict biomass concentration in a fed-batch reactor; the cascaded model was also validated for process control in a closed-loop environment. Statistical process control (SPC) [**Box 3**] is well-established in various industries, working with control charts which indicate whether a process is running in-specification or not. As an example for application of ML, a *long short-term memory network*, which is a variation of RNNs, was used to learn which raw data correlated to important control chart patterns [133].

In addition to MPC and SPC, recent approaches make use of reinforcement learning [134-138]. Petsagkourakis *et al*. [134] applied a Policy Gradient algorithm [**Box 3**] to update a control policy in a batch-to-batch learning approach using true plant data. A surrogate model for the system was used in two simulated case studies to avoid a large number of costly evaluations of the true system for training. In both case studies, the approach outperformed nonlinear MPC, thus posing an interesting starting point for further applications on real plants. In another study using reinforcement learning, Treloar *et al*. [137] applied different variants of the Q-learning algorithm [**Box 3**] to control microbial co-cultures via two different auxotrophic nutrients (**Figure 1D**). Using data from a chemostat model, which was applied for 5 parallel reactors, the authors showed that a control policy could be learned within a 24-hour experiment. For long sample-and-hold intervals, the strategy outperformed a classic PI controller, thus giving a promising outlook for future applications on industrial co-cultures.

While ML is starting to enhance classical process monitoring and control, further applications such as predicting running production costs and attrition rates due to changing resources are still lacking at this stage. Transfer of knowledge, however, can save both time and costs and thus, de-risking can be achieved by reduced trial-and-error approaches [24]. Nonetheless, availability of high-quality data from a significant amount of bioprocess development campaigns is a major challenge since sharing such valuable data in public is not likely to happen. Additionally, full process data is often only available for production runs within specification since a predictably failing run would mean high loss of resources. This results in an imbalance of datasets available for ML [139,140]. Hence, knowledge transfer models seem realistic only as a company-internal project because of very likely non-disclosure of corresponding, valuable data.

**Opportunities for ML in bioprocess development**

ML approaches in bioprocess development show promising results, especially in the areas of strain selection, bioprocess optimization and control. For the first two areas, this is increasingly facilitated by the availability of high-quality data from a wide search space of possible process parameters, easily acquired from established HTS. For bioprocess scale-up, applications and ML models that span various scales are scarce. However, efficient models across scales would have a high impact, particularly by ensuring that HTS data at small-scale are representative for industrial scale. Since the understanding of parameter variation during scale-up is still limited, data-driven ML methods are promising. To address these challenges in the future, transfer learning, i.e., reusing data from other conditions, and increasing well-annotated data should be focused on (see outstanding questions).

Information content of data significantly differs, from highly diverse and broad information at small-scale strain selection to very specific process information at commercial-scale process control. As a consequence, we are convinced that no specific ML method can cover all areas, but rather emphasize the demand for an umbrella of ML methods that can be flexibly combined for each bioprocess. This is also reflected in **Table 1**, which shows that different areas of bioprocess development are covered unequally and that application of ML methods selectively clusters in certain areas. In particular, we found that ANNs are widely used across the bioprocess development fields and that reinforcement learning is dominantly applied in process control. The development of novel ML models in other fields, e.g. physics-informed neural networks, which function as surrogate models for complex mechanistic process models, are currently expanding the available toolbox. Especially in the context of digital twins, which have the purpose to digitally mimic the real system, combining different ML models with mechanistic models for other process modules is promising. This, however, introduces more mathematical and computational challenges for bioprocess modeling, which still need to be tackled.

Nowadays, the biotechnological pipeline is often rather linear, meaning that a funnel exists from early-stage screening to process validation at larger scale, in which the design space is narrowed down after each stage [35]. As a consequence, design iterations are currently mostly taking place at individual stages such as narrowing down the number of strains at small-scale. Here, ML methods help to balance exploration (searching new parameter combinations) vs. exploitation (suggesting the best parameters using the so-far available data) during iterative experimentation. Eventually, they provide processed and abstracted information, enabling the user to make *smarter choices*. Especially at commercial scale, decisions may result in significant cost and resource demands, so that human responsibility and accountability are so far not delegated to a decision-making algorithm.

Ultimately, however, current developments lay the foundation for integrative ML approaches covering more than one stage of the biotechnological pipeline. One example are so-called algorithmic idea generators, i.e., AI algorithms that provide suggestions for biologically and economically feasible bioproducts and robust processes from scratch. This, however, requires efficient technology transfer to bioprocess applications, availability of large, high-quality datasets, e.g., [141] and the commitment to well-curated databases, including detailed metadata of successful as well as failed bioprocess development experiments. Revisiting the DBTL concept, the ideal approach would be an iterative cycle over the whole biotechnological pipeline, meaning that feedback loops between all stages are established. Ultimately, the funnel of the biotechnological pipeline would need to be replaced by early-stage iterations, e.g., between process validation at large-scale and strain selection/bioprocess optimization at small scale. To realize this vision, ML and AI algorithms ultimately need to foster autonomous decision-making instead of replacing modules in the current linear pipeline. Here, we see potential to significantly improve efficiency in bioprocess development and even change the current mode of operation.

**Concluding remarks and future prospects**

In this paper, we reviewed the advent of ML for bioprocess development, where ML methods are increasingly established as a standard in the data analysis toolbox. However, we see potential to transition from individual tools to frameworks that cover the whole process pipeline. At this point, committing to open-source methodology and databases is required for fast progress [142], meaning that a change in mindset is needed to make data and software publicly available. Indeed, this change is happening while corresponding impacts are considered thoroughly [143-149]. These advances will enable unleashing the variety of ML algorithms to better explore the rich amount of data that remains fairly touched nowadays. We do think that a continuous transition towards ML-driven bioprocess development is happening in our discipline. Exciting times lie ahead, giving rise to a new generation of engineers and scientists who can make use of the vast amount of collected yet unanalyzed data, thus generating new strategies for bioprocess development. To both sides, machine learners and bioprocess engineers, we want to express the need for collaboration, expanded networks and joint training to elevate bioprocess development to a new level.

**Box 1 - Machine learning - Paradigms and general challenges**

Artificial intelligence (AI) is a broad field of science with a constantly changing definition [150]. One of the founders of the discipline, John McCarthy, defined it as "the science and engineering of making intelligent machines" (http://www-formal.stanford.edu/jmc/whatisai.pdf). In practice, machine learning (ML) was the most important field of AI in recent years [151]. In contrast to expert systems, which are knowledge-based systems that emulate human decision-making by rules of reasoning [152], the focus of ML is to improve performance based on *training data*, which can be later applied to *test data*.

ML can be further divided into different subfields and central paradigms. A common distinction is between supervised, unsupervised and reinforcement learning. The first one is using *labeled* data during training, meaning that the data provides the expected output. The goal is to learn functions that can describe relationships between input and output variables [153]. In contrast, unsupervised learning uses unlabeled input data; instead of providing labels in training, this branch of methods is focused on identifying patterns that are inherent to data [154]. Finally, reinforcement learning [**Box 2**] uses a different approach. Here, an *agent* acts in an *environment*, where a *policy* is learned to maximize the long-term *reward* [155]. In contrast to labeled data in supervised learning, the input signals for the reward are often delayed and batch-wise optimization is thus frequent [156].

To illustrate several common challenges and principles of ML, a regression task is visualized in **Figure Box 1**. Regression is a typical supervised problem in which the relation between an observed, dependent variable and an independent variable is targeted. Essentially, a function (red) is learned that captures the trajectory (**Figure Box 1A**). Different ML methods assume different functions, e.g., a linear function in linear regression or a Normal distribution in Gaussian process regression [**Box 3**]. The parameters of these functions are learned using labeled data (black dots), in this case response measurements with known input signals. Uncertainty quantification can be used to quantify how certain the estimation is (blue bands), depending on the input signal (**Figure Box 1A**). The more data available, the lower the uncertainty in the parameter estimation. However, if supervised learning is performed on small sets of training data, overfitting is a common problem (**Figure Box 1B**). In this case, the learned function is too specific for the training data and not able to sufficiently adapt for test datasets. In contrast, a function might be too simple to describe complex datasets, which is called underfit. The challenge of generalization is addressed in the field of transfer learning [**Box 2**], where common structures between models are identified to reuse trained models and make efficient use of historic data for new problems [157]. This field is particularly useful since it makes biotechnological processes, which are often small-data problems, accessible to otherwise big-data ML methods.

Overall, ML provides particularly useful methods when datasets are quite large or too complex for human analysis. ML is also beneficial to predict the behavior of biological systems for which domain knowledge is lacking. The current interest and investments in the AI sector foster rapid progress in method development and industry is acknowledging the growing asset of data [24], thus posing a demand for novel technologies. With advances in deep learning, reinforcement learning and transfer learning [158], enabled by higher computational power, improved storage and lower costs [159], bioprocess development is currently at the verge of a new, data-driven era. An overview about the most important methods for this review is given in **Box 3**.

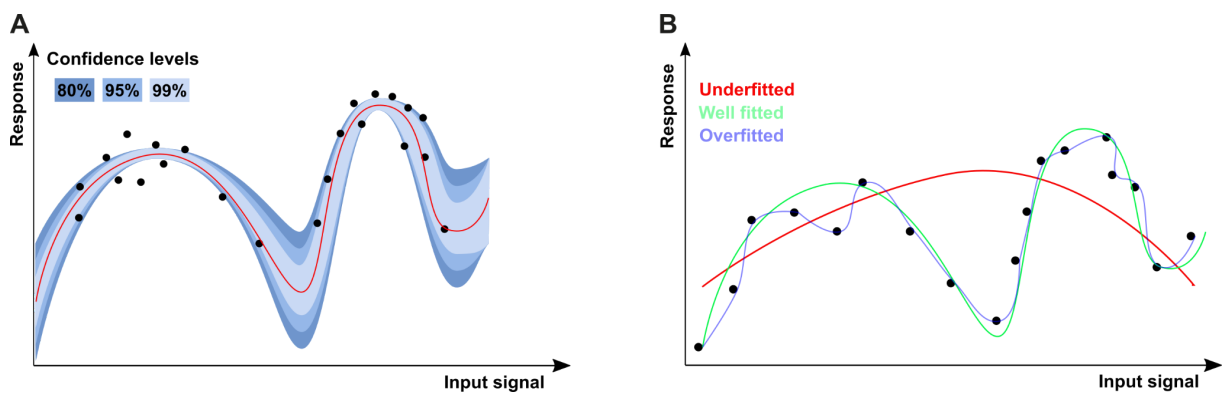

**Figure Box 1: Regression with uncertainty quantification (A) and the overfitting problem (B)**. A: Typical ML task of fitting a model (red) to observed data (black dots) for a response in dependency of the input signal. Uncertainty (blue bands) is lower where more data is observed. B: The number of parameters in a non-linear model indicates its complexity. If it is much smaller than the number of observations, the model often fails to describe the training data (underfit model). If the number of parameters is much bigger than the observations, the model will be unable to generalize beyond the training set. The solution for the underfitting case is straightforward: increase the complexity of the model (number of parameters). In case of a neural network, this could for example mean an additional hidden layer or more neurons per layer. The solution for the overfitting case is reducing the model

parameters. However, if the number of inputs/features is high, it may be impossible to do so and training is computational demanding.

**Box 2 - Machine learning - Overview of concepts**

- **Supervised / unsupervised learning:** See Box 1
- **Deep learning:** This field of ML extracts high-level, often hierarchical, features from raw input by learning how to represent them. Deep neural networks comprise a multitude of layers with (often non-linear) activation functions, whose composition is able to model non-linear dependencies [10,162,163]. Deep learning methods are suitable for complex input such as image data [8].
- **Transfer learning:** A model developed for a specific task is reused as the starting point for a model on a second task. Initial efforts in developing a model structure are not needed anymore for working on the second task. Therefore, less training resources are needed. Key for transfer learning is to then control and handle the corresponding uncertainty [157].
- **Reinforcement Learning (RL):** Trial-and-error approach in which an *agent* acts in an *environment* choosing certain *actions* according to a *policy* that needs to be learned [155]. There are two main approaches: value-based methods and policy-based methods. Value-based approaches try to estimate the value of all actions and states by a function to find the optimal policy; the other type, policy-based algorithms, try to learn the optimal policy directly from a policy space [164].
- **Ensemble learning:** Instead of using a single model to learn, several (different) models are trained on the same dataset. Typically, their combined (or averaged) predictions yield a higher accuracy compared to a single algorithm, but at the cost of higher computing demand [165]. An example are random forests, which are an ensemble of many decision trees (see **Random forests** in **Box 3**).

**Box 3 - Machine learning - Overview of methods and applications**

Machine learning provides researchers with a huge set of methods for data analysis. In the following we will shortly introduce the most important ones. The reader is referred to several recent reviews, which give an overview and short description of ML methods in the fields of biology and biotechnology [160,161]. For details on each method the reader is referred to the included references.

<u>Methods</u>

- **Artificial neural network (ANN):** Network of nodes (neurons), which are connected by edges. Weights on each connection of a neuron influence the output, which is calculated by the weighted inputs. Neurons are typically structured into an input layer, an output layer and hidden layers in-between [166].

- **Recurrent neural network (RNN):** RNNs can operate on ordered data like time-series data or sentences, where the sequence of the individual data points is important. Compared to feedforward neural networks, neurons can share connections and parameters with the same or previous layers. Thereby, information from prior inputs influences current inputs and outputs [167].

- **Convolutional neural network (CNN):** Type of neural networks that is often applied in image processing. The name originates from the additional convolutional layers, which have the purpose to abstract the input by applying filter matrices on it. It is designed to automatically and adaptively learn spatial hierarchies of features, e.g., shapes in hand-written text [168].

- **Bayesian neural network (BNN):** Instead of identifying a single optimal set of parameters to define a neural network, BNNs determine probability distributions for each parameter, thus representing an infinite number of models that can describe the data [169]. This approach allows to quantify the uncertainty in parameters of a neural network [124].

- **Autoencoder:** Autoencoders have the purpose of mapping high-dimensional inputs to a new feature representation (encoding) in a way that the input can be (approximately) reconstructed from the representation [170]. Although variational autoencoders (VAE) share basic architectural features with autoencoders, their purpose and mathematical formulation differ significantly. VAEs use a variational Bayesian model formulation and are applied as generative models, comparable to the application of GANs [171].

- **Random forests:** This technique is an ensemble learning method for classification and regression from many decision trees. The latter are classifying data by continuously splitting it according to certain features [8] that by Thereby, limited prediction power of an individual tree is overcome by the joint prediction from a forest of such trees [172].

- **Support vector machines:** Algorithm that learns by example to assign labels to objects by separating those into two groups. Separation of groups is achieved by a hyperplane that maximizes its distance to the majority of all elements in the respective groups [173,174]. If applied in regression, the technique is referred to as support vector regression (SVR).

- **Gaussian processes (GP):** Are defined as probability distributions over random functions that approximate sets of data points. The name originates from the fact that a subset of the random variables in a GP can always be described by a multivariate Gaussian distribution. GPs are often applied to learn (multi-dimensional) functional relationships from iteratively generated data [175]. Associated uncertainty of predictions can guide iterative experimental design towards experimental conditions with possible high reward.

- **Generative adversarial networks (GANs):** Generative modeling is an unsupervised learning task that involves automatically discovering and learning the regularities or patterns in input data. It is done such that the model can be used to generate or output new examples that plausibly could have been drawn from the original dataset [176].

- **Embedding vectors:** Concept taken from Natural Language Processing (NLP), where it is often referred to as word embedding. The goal in the context of NLP is to quantify similarity of words, e.g., semantic similarity. The method uses vector spaces, thus assigning real numbers to each feature [177]. Modern word embeddings in NLP are based on learning weights of neural networks [178].

- **Policy gradient algorithms:** A policy-based reinforcement learning technique that relies on optimizing parametrized policies with respect to the expected return (long-term cumulative reward) using gradient descent [164].
- **Q-learning algorithm:** A model-free, value-based reinforcement learning algorithm. The Q value refers to the expected reward of playing an action at a certain state following a specific policy [179].

Applications

- **Statistical process control (SPC):** Method of controlling any process based on monitoring temporal evolution of statistics derived from measurements that were demonstrated to be representative for process performance. Based on historical data, specification limits are determined which indicate whether the process runs in-spec or action must be taken, depending on different patterns observed in the control charts.

- **Model predictive control (MPC):** A control concept to predict the future behavior of the controlled system by online assessment of current data. MPC computes an optimal control input while ensuring satisfaction of given system constraints [180]. In contrast to SPC, which is evaluating (validated) limits of summary statistics derived from the process, MPC is using predictive models to forecast the temporal evolution.

- **(Text) data mining:** This ML-powered technology uses natural language processing to examine large volumes of documents for extraction and structuring of contained information. Use cases are, e.g., to discover new information that helps answer research questions [181,182].

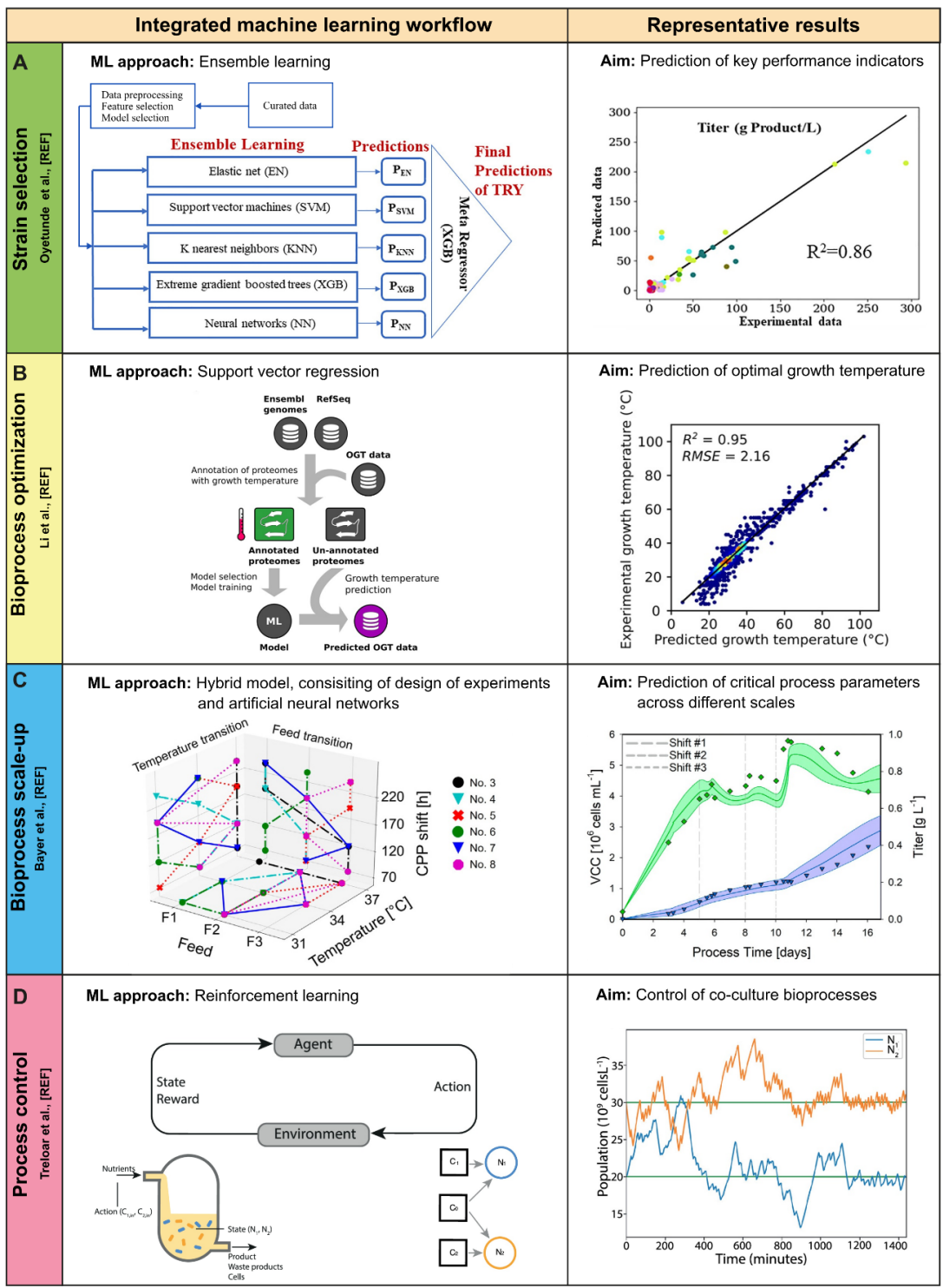

**Figure 1: Overview of key studies of ML Approaches within bioprocessing application for the four fields**

**(A-left)** Machine learning pipeline: Ensemble learning using stacked regressors **(A-right)** Prediction of production metrics on the example of titer. R²: coefficient of determination. Solid lines are shown on the diagonal that represent where all the points would lie for perfect prediction. Colored dots represent different products, ranging from fatty acids to amino acids. Reprinted from [52] with permission.

**(B-left)** Schematic overview of building a machine learning model to predict optimal growth temperature (OGT) for cells. **(B-right)** Performance of the final support vector regression (SVR) model trained on dipeptide data. The correlation between predicted optimal growth temperatures and those present in the original annotated dataset was evaluated. RMSE: root-mean-square error. Colors indicate the density of the points. Reprinted from [64] with permission. Copyright 2019 American Chemical Society.

**(C-left)** Design space of the intensified Design of Experiments (iDoE) performed on the 15 L scale. Critical process parameter transitions for iDoE are displayed as additional planes (z-axis). The individual iDoE bioprocesses are represented by different colors and symbols. **(C-right)** Prediction of large-scale critical process parameters in an iDoE with a hybrid model trained on data from shaker-scale experiments. The model estimations for the viable cell concentration (VCC, green lines) and product titer (blue lines) are indicated along with the respective confidence interval (shaded area). The analytical measurements are given for the VCC (green squares) and the product titer (blue triangles). Reprinted from [101] with permission.

**(D-left)** Machine learning pipeline using reinforcement learning [**Box 2**] for the control of bioprocess co-cultures in a simulated chemostat. The agent adds nutrient sources C1 and C2 to control the co-culture composition. **(D-right)** Co-culture population in a chemostat. During the exploration phase, in which the agent learns the policy, the population levels (N1; N2) vary and random actions are taken; as the exploration rate decreases (shift towards exploitation), they move to the target values (green lines). Reprinted from [137] with permission.

**Table 1 - Application of machine learning algorithms during bioprocess development**

| ML algorithm | Stage | | | |
|---|---|---|---|---|
| | **Strain selection** | **Bioprocess optimization** | **Scale-up** | **Process control** |
| ANN | | [68-73,85,183-190] | [101,191] | [118-120,122,131,137,192,193] |
| RNN | [54] | | [99] | [133,134,194] |
| CNN | [52,195] | [84] | | |
| Unsupervised Feature Representation (e.g., autoencoders) | | | | [122,123] |
| Trees / Random forests | [52] | [196,197] | [100] | [198] |
| SVM / SVR | [52] | [64,87] | [97] | [108,120,199] |
| Gaussian Processes | | [65-67,75] | | [132] |
| GANs / Variational Autoencoders | [61] | [83] | | |
| Graph-based neural networks | | [91] | | |
| Transfer Learning | [200] | [74,75,201] | | [127] |
| Reinforcement Learning | [62] | [91,202] | | [134-138,203] |
| Ensemble learning | [52,53,57,204,205] | [206-208] | | [209,210] |
| Text data mining | [51,211] | [212,213] | | |

**Highlights**

- Bioprocess development requires identification of robust design spaces for specific bioproducts and involves efficient strain selection, bioprocess optimization, scale-up and optimal control strategies for robust industrial production.
- Beyond multivariate data analysis, deep learning, reinforcement learning and other novel ML techniques start to complement and replace traditional data analysis approaches to accelerate screening, optimization and control procedures.
- Transfer learning is emerging as a means to leverage the potential of historic data to guide novel production processes.
- No single algorithmic solution will be suitable for all aspects of bioprocess development. Instead, a flexible combination of various techniques is required to enhance the whole development pipeline.
- Fast impact is expected in autonomous strain selection and the optimization of bioprocess parameters. The application of ML for scale-up has a high impact but needs further development.

**Outstanding questions**

- In the current biotechnological pipeline, strain selection and optimization of physico-chemical parameters is often performed sequentially. These stages are, however, highly interdependent. Can ML induce a shift in the conventional, linear pipeline towards an integrative, circular approach, including feedback mechanisms between the two stages?
- Can proof-of-concept ML studies that have been conducted for well-established bioprocesses be transferred to (non-model) production hosts and bioprocesses?
- To realize the above-mentioned goals, thorough data collection and annotation with metadata, corresponding data formats and databases, interfaces for automation and other technical requirements need to be established. How can we realize and motivate this change? Are the FAIR data principles [214] sufficient and how can this transition be implemented in the mindset of bioprocess engineers and data scientists?

- Publishing and using negative results as training data can be game-changing for improved ML models in bioprocessing. How can we motivate and reward publication of negative results?
- Which data is more suitable for different ML models: sparse and highly informative data that is expensive to acquire (e.g. infrequent mass spectrometry data) vs. big amounts of data that are less informative but can be retrieved more cheaply (e.g. online spectroscopy)?
- Uncertainty quantification of ML results and interpretability of ML procedures are important for assessing model quality and revealing biological meaning/ drawing meaningful conclusions. How can we foster the integration of uncertainty quantification in biotechnology to increase our process understanding?

**Glossary**

**Backpropagation:** In the context of artificial neural networks, backpropagation refers to a method that allows to calculate the gradient of the loss function, which is used for training the network, i.e. identifying its weights using data [170].

**Bayesian statistics:** Field of statistics covering methods with a Bayesian interpretation of probability, which according to Bayes theorem includes prior beliefs in an event. Methods cover, among others, Bayesian inference of parameters, statistical modeling, Bayesian optimization (for sequential design) and Bayesian networks.

**Bioprocess development:** Bioprocess development aims for the identification of a robust design space for the production of a specific bioproduct with desired yield and purity. It requires experiments and data analysis to understand the interaction of parameters within the specific bioprocess.

**Constraint-based modeling (COBRA):** Subfield of systems biology that takes into account the underlying physical, enzymatic, and topological constraints of a phenotype in a metabolic network [215]. Methods include Flux Balance Analysis (FBA) [46], Minimization of Metabolic Adjustment (MOMA) [47], or Minimal Cut Sets (MCS) [48].

**Critical process parameters (CPP):** Parameters of a (bio-)pharmaceutical production process that have been shown to affect the critical quality attributes of the final product. The parameter values have to be monitored and to be kept in proven ranges to not affect the corresponding critical quality attributes in a negative way.

**Critical quality attributes (CQA):** Attributes of a (bio-)pharmaceutical product that determine its quality and for which certain value ranges have to be met in order to release the product.

**Design-Build-Test-Learn (DBTL) cycle**: A loop used recursively to obtain a design that satisfies the desired specifications. In bioengineering, the DBTL cycle makes use of synthetic biology to engineer biomanufacturing solutions for industrial application.

**(Intensified) Design of Experiments (iDoE / DoE):** In Design of Experiments, the goal is to create an empirical model that describes how a process responds to changes in influential factors. It is often performed in two stages: 1. screening for identification of influential factors and 2. prediction of response surfaces to identify optimal operation conditions [216]. Intensified DoE is an adaptation in which several set points for influential factors are tested as intra-experiment variations, thus reducing the overall number of experiments [217].

**Digital twins:** Digital twins are detailed, virtual representations of production systems, where feedback between model and physical systems is characteristic [218]. Applications include real-time monitoring of manufacturing processes and fault detection [116].

**Flux Balance Analysis (FBA)**: A mathematical method to solve stoichiometric metabolic networks for steady state flux solutions. Applications include identification of targets for metabolic engineering and media design.

**Hybrid modeling:** Combines data-driven models with mechanistic, *a priori* knowledge into one superior model structure.

**Linear regression**: A linear approach for modeling the relationship between a response and one or more explanatory variables.

**Natural Language Processing:** Field of AI that is concerned with automatically analyzing and representing human language; as such it is closely related to computer science and linguistics [219]. Applications include speech recognition, machine translation and synthesis of language [220].

**Non-linear regression:** A form of regression analysis in which observational data are modeled by a function which is a nonlinear combination of the model parameters and depends on one or more independent variables.

**Process analytical technology:** System for designing, analyzing and controlling (pharmaceutical) manufacturing processes through measurements of critical quality and performance attributes (https://www.fda.gov/media/71012/download).

**Surrogate modeling**: Surrogate models are approximations that are used when the desired output of a system is expensive or hard to simulate [2]. An example are physics-informed neural networks (PINNs) that can for example be used to replace costly simulations in computational fluid dynamics [102].